\documentclass[letterpaper, 10 pt, journal, twoside]{IEEEtran}  %

\IEEEoverridecommandlockouts                              %

\usepackage{amssymb}  %
\usepackage{amsmath}
\usepackage[table]{xcolor}
\usepackage{graphicx}
\usepackage[T1]{fontenc}   %
\usepackage{ccicons}          %
\usepackage{paralist}
\usepackage{hyperref}
\usepackage[nolist]{acronym}
\usepackage{graphics}
\usepackage{svg}
\usepackage{colortbl}
\usepackage{hhline}
\usepackage{makecell}
\usepackage{footnote}

\usepackage{color}

\newtheorem{definition}{Definition}
\newtheorem{example}{Example}

\newcommand{\comment}[1]{\ignorespaces}
\newcommand{\maybecite}[1]{\ignorespaces}
\newcommand{\quotes}[1]{`#1'}

\newcommand{\edit}[1]{#1}

\title{Explanation-Aware Experience Replay in Rule-Dense Environments}

\author{Francesco Sovrano$^{1,*}$, Alex Raymond$^{2,*}$, and Amanda Prorok$^{2}$%
\thanks{}
\thanks{Alex Raymond is supported by the Royal Commission for the Exhibition of 1851 and L3Harris ASV. Amanda Prorok is supported by the European Research Council (ERC) Project 949940 (gAIa). Their support is gratefully acknowledged.}
\thanks{$^{1}$Francesco Sovrano is with the Department of Computer Science, University of Bologna, Italy
         {\tt\small francesco.sovrano2@unibo.it}}%
\thanks{$^{2}$Alex Raymond and Amanda Prorok are with the Department of Computer Science and Technology, University of Cambridge, United Kingdom
         {\tt\small alex.raymond@cl.cam.ac.uk, asp45@cam.ac.uk}}%
\thanks{*Authors contributed equally to this work and are listed as co-first authors.}
}

\begin{document}

\maketitle

\begin{abstract}

Human environments are often regulated by explicit and complex rulesets. Integrating Reinforcement Learning (RL) agents into such environments motivates the development of learning mechanisms that perform well in rule-dense and exception-ridden environments such as autonomous driving on regulated roads. In this paper, we propose a method for organising experience by means of partitioning the experience buffer into clusters labelled on a per-explanation basis. We present discrete and continuous navigation environments compatible with modular rulesets and 9 learning tasks. For environments with explainable rulesets, we convert rule-based explanations into case-based explanations by allocating state-transitions into clusters labelled with explanations. This allows us to sample experiences in a curricular and task-oriented manner, focusing on the rarity, importance, and meaning of events. We label this concept Explanation-Awareness (XA). We perform XA experience replay (XAER) with intra and inter-cluster prioritisation, and introduce XA-compatible versions of DQN, TD3, and SAC. Performance is consistently superior with XA versions of those algorithms, compared to traditional Prioritised Experience Replay baselines, indicating that explanation engineering can be used in lieu of reward engineering for environments with explainable features.

\end{abstract}

\section{INTRODUCTION}

\IEEEPARstart{H}{uman-regulated} environments often rely on legislation and complex sets of rules. At present, Reinforcement Learning (RL) methods are usually tested in environments with relatively sparse rules and exceptions \cite{Kaiser2019Model-basedAtari}. 
Denser regulations appear in applications of RL for autonomous vehicles research, but such rulesets are often fixed in terms of complexity  \cite{Li2019UrbanLearning}. 
We are interested in observing how learning can be affected by the depth of the rulesets governing such systems. 
With large numbers of corner cases arising as a consequence of dense rulesets, generating a sufficiently diverse set of experiences and exposing these exceptions to an RL agent can be challenging.
Some works in literature propose to sample past experiences related to those exceptions, heuristically revisiting potentially important events. 
Among them, the technique of Prioritised Experience Replay (PER) \cite{Schaul2015PrioritizedReplay} looks at over-sampling experiences that are most poorly captured by the agent’s learned model. However, this mechanism does not necessarily focus on the cause of events or their exceptional nature.

In this letter, we pursue the intuition that explanations are a pivotal mechanism for human intelligence, and that this mechanism 
has the potential to boost the performance of RL agents in complex environments. 
This is why we draw inspiration from user-centred explanatory processes for humans \cite{sovrano2021philosophy}, and design a set of heuristics and mechanisms for prioritised experience replay to explain complex regulations to a generic off-policy RL agent.
A central design challenge towards this goal is integrating explanations into computational representations. Approaches such as encoding the ruleset (or part of it) into the agent's observation space may incur severe re-training overhead even under minimal ruleset changes, as the semantics of the regulation are explicitly provided as input \cite{Kiran2021DeepSurvey}. 
This minimises compatibility with extant methods and may obscure whether differences in performance are due to changes to the architecture or the complexity of the ruleset. 
We propose a solution that is agnostic to explicitly engineering state and observation spaces, using an explanation-aware experience replay mechanism.

\begin{figure}[!htb]
	\centering
	\includegraphics[width=\columnwidth]{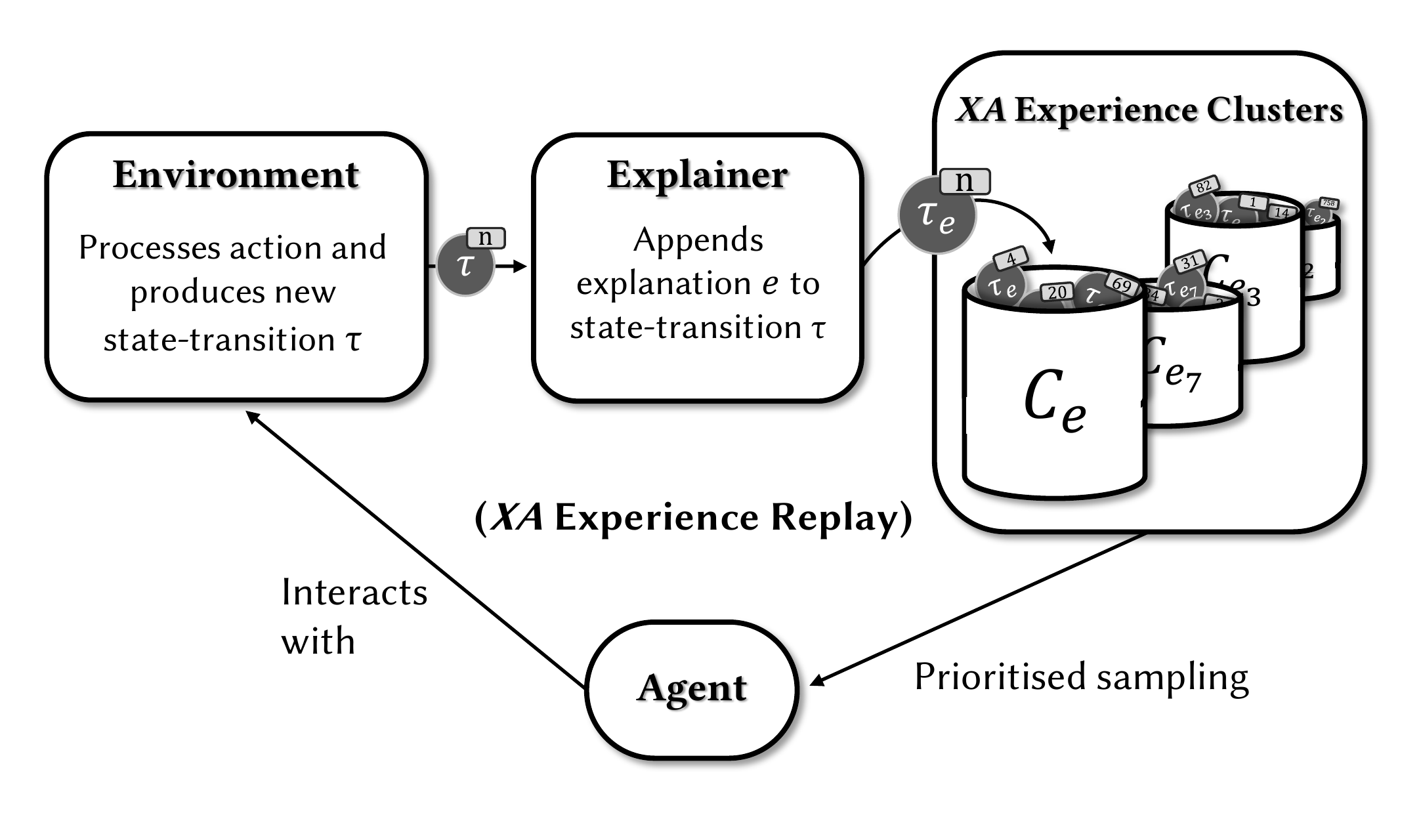}
	\caption{Overview of XAER. The explainer labels a state-transition $\tau$ with an explanation $e$, which is stored in a cluster ($C_e$) containing other experiences labelled with the same explanation.}
	\label{fig:diagram1}
\end{figure}

In our approach, we avoid explicit representations of the ruleset (i.e. rule-based explanations \cite{Branting1991BuildingCases}) by instead representing the meaning of the regulations \textit{as organised collections of examples} (i.e. case-based explanations \cite{Aamodt1994Case-BasedApproaches}). 
These explanations do not need to be \textit{understood} by the agent in the traditional sense, but can still convey meaning if the example was labelled/explained in a semantic and meaningful process. 
In a ludic example, suppose a young man, called Luke, is taking hyperspace flight lessons from his exasperated friend Chewbacca. However, he does not understand a single word of Shyriiwook, the tutor's language. With sufficient repetition, Luke can associate distinct Wookiee growls (and punishments) to categories of experienced episodes, even if the content of the message is in an unknown language. Eventually, Luke would learn the meaning of the most relevant utterances by associating them to the experienced consequences.
Hence, our approach modifies conventional experience replay structures by partitioning the replay buffer (or memory) into multiple clusters, each representing a distinct explanation associated with a collection of experiences that serve as examples.
We call this process \textit{Explanation-Aware Experience Replay (XAER)} (see Figure~\ref{fig:diagram1}) and integrate this technique into three seminal learning algorithms: Deep Q-Networks (DQN) \cite{Mnih2015Human-levelLearning}, Twin-Delayed DDPG (TD3) \cite{Fujimoto2018AddressingMethods}, and Soft Actor-Critic (SAC) \cite{Haarnoja2018SoftActor}. 

In summary, we state the following contributions:
\begin{itemize}
    \item We show how distinct types and instances of explanations can be used to partition replay buffers and improve the rule coverage of sampled experiences.
    \item We design discrete and continuous environments (GridDrive and GraphDrive) compatible with modular rulesets of arbitrary complexity (cultures). This leads to 9 \textit{learning tasks} involving both environments with different levels of rule complexity and reward sparsity. These serve as a platform to evaluate how RL agents react to changes in rulesets whilst keeping a consistent state and action space. 
    \item We introduce XAER-modified versions of traditional algorithms such as DQN, TD3, and SAC, and test the performance of those modified versions in our proposed environments.
\end{itemize}
Upon experimenting on the proposed continuous and discrete environments, our key insight is that organising experiences with XAER improves agent performance (compared to traditional PER) and can be able to reach a better policy where traditional PER may fail to learn altogether.

\section{RELATED WORK} \label{sec:background}
In this section we give the necessary background to understand our proposed solution and the following experiments, along with prior related work. %

\subsection{Model-Free Reinforcement Learning}
A Reinforcement Learning problem is typically formalised as a Markov Decision Process (MDP). In this setting, an agent interacts at discrete time steps with an external environment. At each time step $t$,  the agent observes a state $s_t$ and chooses an action $a_t$ according to some policy $\pi$, that is a mapping (a probability distribution) from states to actions. As a result of its action, the agent obtains a reward $r_t$, and the environment passes to a new state $s'=s_{t+1}$.
The process is then iterated until a terminal state is reached.

The future cumulative reward $R_t = \sum^\infty_{k=0}\gamma^{k}r_{t+k}$ is the total accumulated reward from time starting at $t$. $\gamma \in \left[ 0, 1 \right]$ is the {\em discount factor}, representing the difference in importance between present and future rewards.
The goal of the agent is to maximise the expected cumulative return starting from an initial state $s=s_t$. 
The \textit{action  value} $Q^\pi(s,a)  = E^{\pi}[R_t|s=s_t,a=a_t]$ is the expected return for selecting action $a$ in state $s_t$ and prosecuting with strategy $\pi$. 
Given a state $s$ and an action $a$, the optimal {\em action value} function $Q^*(s,a) = \max_\pi Q^\pi(s,a)$ is the best possible action value achievable by any policy. 
Similarly, the \textit{value} of state $s$ given a policy $\pi$ is $V^\pi(s)  = E^\pi[R_t|s=s_t]$ and the optimal value function is $V^*(s) = \max_\pi V^\pi(s)$.

Two of the major approaches to RL are value-based and actor-critic algorithms. Value-based algorithms, such as Deep Q-Networks (DQN) \cite{Mnih2015Human-levelLearning}, use temporal difference learning, where policy extraction is done after an optimal value function is found. 
Actor-critic methods, such as and Twin-Delayed DDPG (TD3) \cite{Fujimoto2018AddressingMethods} and Soft Actor-Critic (SAC) \cite{Haarnoja2018SoftActor}, rely on evaluating and improving a policy (via gradient descent) together with a state-value function.
DQN is one of the very first value-based deep RL algorithms, designed to work on discrete action-spaces only. %
Adaptations for continuous action-spaces, as DDPG and then TD3, propose to address value overestimation problems by means of clipped double Q-learning, delayed update of target and policy networks, and target policy smoothing. 
However, one of the main limitations with TD3 is that it randomly samples actions using a pre-defined distribution.
To overcome the issue of being limited by a fixed distribution, Soft Actor Critic (SAC) empowers the agent with the ability to also learn the distribution with which to sample actions, empowering the agent to explore more different strategies through entropy maximisation. 

\subsection{Explanations in RL}
The most important field studying explanations in AI and RL is eXplainable AI (XAI) \cite{BarredoArrieta2020ExplainableAI}. 
Among the many surveys on XAI, a common dimension used to classify explanations is the representative format used to convey them.
Within this domain, explanations are commonly conveyed via textual/visual descriptive representations of the decision criteria (i.e. \textit{rule-based}), or with similar examples (i.e. \textit{case-based}). %
An example of rule-based explanation is \textit{`you will get a penalty for reaching 75, which is above the speed limit of 50'}, based on the rule \textit{`if speed is above 50, you will get a penalty'}.
While an example of case-based explanation is \textit{`you get a penalty because you are in a situation similar to this other vehicle that reached speed 74 and was previously penalised'}.

Dietterich and Flann \cite{dietterich1997explanation} frame explanation-based RL as a case-based explanatory process where prototypical trajectories of state-transitions are used to tackle similar but unseen situations, while Chow et al. \cite{chow2018lyapunov} implement a rule-based method, constraining the Markov Decision Process by means of Lyapunov functions. %

Generally speaking, many rule-based methods for explaining to RL agents usually fall under the umbrella of a sub-discipline called Safe RL \cite{garcia2015comprehensive}. 
Safe RL includes techniques for both: encoding rules in the optimality criterion \cite{chow2018lyapunov, Rong2020SafeDriving} and incorporating such external knowledge into the action/state space \cite{Berkenkamp2017SafeGuarantees}. 
Although not generating explicit explanations, those methods engineer safety rules into the learning process, implicitly explaining to the agent what \textit{not} to do. 
Alternatively, a famous example of case-based methods for explaining to RL agents is that of Imitation Learning \cite{hussein2017imitation}, where demonstrations (as trajectories of state-transitions generated by a human or expert algorithms) are used to train the RL agent. These can be seen as high-quality cases/examples provided by an expert human or algorithm. However, access to human expert data may not scale well to every domain, and not all problems dispose of accessible expert algorithms.

We are interested in sampling the most useful experiences to cover a particular agent's gap in knowledge. An \textit{agent-centred} explanatory process is an iterative process that follows the agent through the process of learning and selects the most useful explanations for it, at every time-step. Below, we look at how experience replay techniques tackle this issue in off-policy RL.

\subsection{Prioritised Experience Replay}
Algorithms such as DQN, TD3, and SAC aim to find a policy that maximises the cumulative return, by keeping and learning from a set of expected returns estimates for some past policy $\pi$. This set of expected returns is kept in an experience buffer, enabling \textit{experience replay}.
\textit{Experience replay} \cite{Schaul2015PrioritizedReplay} consists in re-utilising information from the space of sampled experiences. The agent's experiences at each time-step $t$ are stored as transitions $e_{t} = \left(s_{t}, a_{t}, r_{t}, s_{t+1}\right)$, where $s_t$, $a_t$, $r_t$ represent the state, action, and reward at time $t$, followed by the next state $s_{t+1}$. These transitions are pooled over many episodes into a replay memory, which is usually randomly sampled for a mini-batch of experiences. %

Experience sampling can be improved by differentiating important transitions from unimportant ones. In Prioritised Experience Replay (PER) \cite{Schaul2015PrioritizedReplay}, the importance of transitions with high expected learning value is measured by the magnitude of their temporal-difference (TD) error. Experiences with larger TD are sampled more frequently, as TD quantifies the unexpectedness of a given transition \cite{Li2021RevisitingPerspective}. This prioritisation can lead to a loss of diversity and introduce biases. 
Bias in prioritised experience replay occurs when the distribution is changed without control. This effect therefore changes the solution that the estimates will converge to. This bias can be corrected through importance-sampling (IS) weights.

Many approaches to Prioritised Experience Replay (PER) in RL \cite{Schaul2015PrioritizedReplay} can be re-framed as mechanisms for achieving agent-centrality, re-ordering experience by relevance in the attempt of explaining to the agent and selecting the most useful experience, as indirectly suggested by Li et al. \cite{Li2021RevisitingPerspective}.
Over the years, many human-inspired intuitions behind PER drove researchers towards improved, more sophisticated and agent-centred mechanisms to RL \cite{sun2020attentive, yin2017knowledge, zha2019experience}.
Among these works, the closest to a fully agent-centred explanatory process is Experience Replay Optimisation \cite{zha2019experience}, which moves towards agent-centrality by providing an external black-box mechanism (or experience sampler) for extracting arbitrary sequences of information out of a flat (no abstraction involved) experience buffer. The experience sampler is trained to select the most `useful' ones for the learning agent. However, due to its non-explainable nature, it is not clear whether the benefits given by Experience Replay Optimisation are due to the overhead given by the experience sampler increasing the number of neurons in the agent's network.

Another work trying to achieve agent-centrality in this sense is Attentive Experience Replay \cite{sun2020attentive}, suggesting to prioritise uncommon experience that is also on-distribution (related to the agent's current task). However this work, as the previous one, also falls short of explicitly organising experience in an abstract-enough way by conveying human-readable explanations to the agent. Hierarchical Experience Replay \cite{yin2017knowledge} has attempted to address the abstraction issue in an attempt to simplify the task to the agent, decomposing it into sub-tasks. However, they do not do so in an agent-centred and goal-oriented way, given that its sub-task selection is uniform and not curricular.

On the other hand, a curricular approach for training RL agents was proposed by Ren et al. \cite{ren2018self}, exploiting PER and the intuition that simplicity is inversely proportional to TD-errors, but not exploiting any abstract and hierarchical representation of tasks.
Similarly to ours, \cite{sovrano2019combining} aims to organise experience abstractly, based on its explanatory content --- framed as the ability to answer how good/bad a sequence of state-transitions is with respect to average experience. This work only considers explanations about the immediate performance of the agent (i.e. \texttt{HOW} explanations), and lacks any consideration of other and richer types (i.e. \texttt{WHY}), as well as curricular prioritisation facilities.

\section{EXPLANATION-AWARENESS}
\label{sec:xa}

Our use of explanations is aligned to Holland's \cite{holland1989induction} and Achinstein's \cite{achinstein1983nature} philosophical theories of explanations. 
In fact, in the former, the act of explaining is framed as a process of revising belief whenever new experience challenges it. In the latter, explaining is the attempt to answer questions (such as `why', `what', etc \cite{sovrano2021philosophy}) in an agent-centred way.
Specifically, we propose a transformation of rule-based explanations (e.g. given by a ruleset/culture) to case-based explanations (experience), which are compatible with experience replay. 
Leaning on the concept of \textit{Explanation-Awareness (XA)}, our heuristics facilitate information acquisition via the organisation of experience buffers. 

Drawing from an epistemic \cite{mayes2005theories} interpretation of explanations, we argue that a central aspect of providing case-based explanations to an RL agent comes from meaningfully re-ordering experience to a greater degree. 
The intuition behind how we construct our case-based explanations is: \textit{\quotes{a \underline{simple} set of \underline{relevant} state-transitions representing \underline{abstract}-enough aspects of the problem to be solved.}}
This intuition motivates the heuristics of \textit{abstraction}, \textit{relevance}, and \textit{simplicity} (ARS, in short). 
We adapt these heuristics from prior work \cite{sovrano2021philosophy} in the HCI domain, where they are presented in greater abstraction to form a higher-level taxonomy and knowledge graph for an interactive explanatory process.

Consider a problem where an RL agent has to learn a policy to optimally navigate through an environment with sophisticated rules and exceptions (e.g. a real traffic regulation with exceptions for special types of vehicles). Let the state-transition $\tau = (s_t, a_t, r_t, s_{t+1})$ denote the transition from state $s_t$ to state $s_{t+1}$ by means of action $a_t$, yielding a reward $r_t$. 
We assume the environment is imbued with explanatory capabilities via an \textit{explainer}. Note that the explanations generated by the explainer can have virtually any representation, be it human-understandable or not, provided they are distinct and serve the purpose of labelling different clusters.

\begin{definition}[Explainer]
The explainer $\epsilon: \Omega \rightarrow ES$ is a function that maps a list of state-transition tuples $\tau \in \Omega$ to an explanation $e_{r} \in ES$, where $\Omega$ is the space of possible state-transitions and $ES$ is the explanatory space, i.e., the space of all possible explanations. %
\end{definition}

An agent who has more diverse experiences with regards to the \textit{reasons} (explanations) associated with rewards will have a better chance at converging towards a policy that better represents the underlying ruleset. Therefore, we posit that the more complex the environment is in terms of rules, the more useful \textit{Explanation-Awareness (XA)} should be, as it would ensure a more even distribution of experiences with regards to different reasons justifying rewards. This diversity of explanations culminates on a clustering that is semantic by nature, and transitions are partitioned according to the explanation that represented its reward.

\begin{definition}[XA Clusters]
Let $\tau_{e} = (s_t, a_t, r_t, e_{r_t}, s_{t+1})$ be a XA state-transition represented by the explanation $e$, where \edit{$\tau_e: \tau \times e_r$, $\tau \in \Omega$ and $e \in ES$}. Let $\Omega$ be the set of all state-transitions. We say \edit{$\mathcal{C} = \{C_{e_1}, \dots, C_{e_k}\}$} is the set of XA clusters seen in $\Omega$, where $k$ is the number of different explanations seen. 
\end{definition}

We introduce our adaptation of ARS, below.

\subsection{Abstraction: Clustering Strategies}
The purpose of the \textit{abstraction} heuristic is to regulate the level of granularity of the explanations, hence of the experience clusters.
Our abstractions are based on the understanding that explanations are indeed answers to questions.
Hence, explanations may have different granularity defined by the level-of-detail of the question they answer.

More in detail, the \texttt{HOW} explanations we consider answer the question \quotes{How well is the agent performing with this reward?}. 
This type of explanations can be produced by studying the average behaviour of an agent. For example, if an episode has a cumulative reward that is greater than the running mean, then the explanation indicates that the agent is behaving better than average. \edit{Hence, these \texttt{HOW} explanations do not need to be designed with any specific domain knowledge, as they are governed exclusively by the performance of the agent.}
On the other hand, the \texttt{WHY} explanations we consider answer the question \quotes{Why did the agent achieve this reward?}. \edit{These \texttt{WHY} explanations could depend on an explainer function with task/domain knowledge that can distinguish and cluster types of transitions (see Example~\ref{example:football}, below).}
Furthermore, \texttt{WHY} and \texttt{HOW} explanations (or any other type) can be combined so that the explanation would answer both the associated questions. %

In order to compose the experience buffer, represented by the set of experience clusters \edit{$\mathcal{C} = \{C_{e_1}, \dots, C_{e_k}\}$}, we consequently devise the following clustering strategies, for each explanation type:
\begin{enumerate}
    \item \texttt{HOW}: The experience buffer is divided into 2 clusters $C_{\text{better}}$ and $C_{\text{worse}}$, where $C_{\text{better}}$ contains batches with rewards greater than the running mean of rewards, and vice-versa (given a sliding window of a defined size).
    \item \texttt{WHY}: The number of clusters is equivalent to the number of distinct explanations available. If a batch can be explained by multiple explanations simultaneously, we select the explanation associated with the smallest cluster (most under-represented) and the batch is associated to the corresponding cluster.\footnote{Since buffers will be prioritised and clusters will be fairly represented, there is no need for duplicating the batch across multiple clusters.}
    \item \texttt{HOW+WHY}: a combination of \texttt{HOW} and \texttt{WHY} strategies. There are two custom $C_{\text{better}}$ and $C_{\text{worse}}$ clusters for every \texttt{WHY} explanation, formed after their concatenation. 
\end{enumerate}

\edit{\begin{example}
Suppose a hypothetical football environment with a \texttt{WHY} explainer function. This function could either be part of the environment (a logical mechanism that recognises when certain states are reached and produces a state label), or an external mechanism that receives state-transitions as input and produces explanations. The explanations could be generated by the rules of the game, such as `goal', `offside', or `foul'. The corresponding \texttt{WHY} clusters would be $\mathcal{C} = \{C_{\text{goal}}, C_{\text{offside}}, C_{\text{foul}}, \ldots\}$, where each cluster would contain a set of state-transitions associated with each label. If \texttt{HOW+WHY} were used, clusters would be $\mathcal{C} = \{C_{\text{goal\_better}}, C_{\text{goal\_worse}}, C_{\text{offside\_better}}, \ldots\}$.
\label{example:football}
\end{example}}

After clustering state-transitions using the prior clustering strategies, we propose mechanisms for assessing the \textit{relevance} of specific state-transitions during learning.

\subsection{Relevance: Intra-Cluster Prioritisation} \label{sec:relevance_heuristic}
Prioritisation mechanisms are used for organising information given their relevance to the agent's objectives.

The priority of a batch is usually estimated by computing its loss with respect to the agent's objective \cite{Schaul2015PrioritizedReplay}. In DQN, TD3, and SAC, relevance is estimated by the absolute TD-error of the agent. The closer to 0, the lower the loss and the relevance. The intuition is that batches with TD-error equal to zero are of no use since they represent an already solved challenge. In our method, this \textit{relevance} heuristic can be combined with the aforementioned clustering strategy by sampling clusters in a prioritised way (by summing the priorities of all its batches) and then performing prioritised sampling of batches from the sampled cluster.

\subsection{Simplicity: (Curricular) Inter-Cluster Prioritisation}
Occam's Razor \cite{Blumer1987OccamsRazor} states that when presented with two explanations for the same phenomenon, the simplest explanation should be preferred.
In human explanations, \textit{simplicity} is a common heuristic \cite{johnson2019simplicity,nielsen1994enhancing}.
We will adhere to those principles and select minimal and simple explanations, following a curricular approach.

Clustered prioritised experience replay changes the real distribution of tasks by means of over-sampling. %
Assuming that the whole experience buffer has a fixed and constant size $N$, and that the experience buffer contains \edit{$|\mathcal{C}|$} different clusters, let $S_\text{min}$ and $S_\text{max}$ be the minimum and maximum size of a cluster. Any new experience is added to a full buffer by removing the oldest one within buffers having more elements than $S_\text{min}$.

If all the clusters have the same size (therefore $S_\text{min} = S_\text{max}$), replaying the task's cluster with the highest (TD-error) priority might push the agent to tackle the exceptions before the most common tasks, preventing the agent from learning an optimal policy faster. The assumption here is that exceptional tasks (exceptions) are less frequent.

On the other hand, if $S_\text{min} = 0$ and $S_\text{max} = \infty$, the size of a cluster would depend only on the real distribution of tasks within a small sliding window, as in traditional PER, thus preventing over-sampling. The presence of clusters helps over-sampling batches likely related to under-represented tasks, and learning to tackle potentially hard cases more efficiently.

Consequently, we posit that $S_\text{min}$ shall be large enough for effective over-sampling, while having $S_\text{max} > S_\text{min}$ being dependent on the real distribution of tasks. This will push the agent towards tackling the most frequent and relevant tasks first, analogously to curricular learning. We define a hyperparameter to control the \textit{cluster size proportion}.

\begin{definition} [Cluster Size Proportion]
In order for all clusters to have a size $S_\text{min} \leq S \leq S_\text{max}$, we set \edit{$S_\text{max} = S_\text{min} + (\xi-1) \cdot |\mathcal{C}| \cdot S_\text{min}$}, where $\xi \geq 1$ represents the \textit{cluster size proportion}. 
\end{definition}
Therefore, \edit{$S_\text{min} = \frac{N}{|\mathcal{C}| \cdot \xi}$} can be easily controlled by modifying $\xi$. We enforce $S_\text{min} < S_\text{max}$ when $\xi > 1$. 
Consequently, for \textit{curricular prioritisation}, if the cluster’s priority is (for example) computed as the sum of the priorities of its batch, and $\xi > 1$ is not too large (e.g. $\xi = 5$), the resulting cluster’s priorities will reflect the real distribution of tasks while smoothly over-sampling the most relevant tasks. This avoids over-estimation of the priority of a task. \edit{As $\xi$ gives us control of the degree of on-policyness, different values of $\xi$ might perform better with on an algorithm and environment basis\footnote{\edit{However, tuning for $\xi$ seems relatively simple, and a grid search on $\xi \in \{1, 2, 3, 4, 5, \inf\}$ might suffice for most cases.}}. Higher values of $\xi$ mean that the distribution of state-transitions reflects more transitions seen within the current policy, thus being advantageous for entropy-maximisation algorithms such as SAC \cite{Li2021RevisitingPerspective}. Likewise, fully off-policy algorithms such as DQN may exhibit superior results with low values of $\xi$ (e.g. $\xi = 1$).}

With those mechanisms in place, we propose new environments to evaluate the performance of agents when subjected to complex rulesets.

\subsection{Annealing the Bias}
\edit{
Similarly to PER \cite{Schaul2015PrioritizedReplay}, sampling state-transitions from prioritised clusters might produce unwanted bias.
The standard debiasing function of PER weighs expected values using the normalised weight $\frac{P(\bar{\tau})}{P(\tau)} \in [0,1]$, where $P(\tau)$ is the probability of sampling a state transition $\tau$ from the whole buffer and $\bar{\tau}$ is the state-transition with the lowest probability for the whole buffer. 
We adapted the debiasing function of PER by changing the formula to consider the fact that state-transitions are sampled from clusters (which are in turn sampled).
Therefore, the debiasing function of XAER computes the joint probability of sampling both a cluster $c$ and a state-transition $\tau$.
More precisely, considering that the two events are not independent, we compute this joint probability as $P(c) \cdot P(\tau|c)$.
Hence, the normalised weights produced by the debiasing function of XAER are given by $\frac{P(\bar{c} \cap \bar{\tau})}{P(c \cap \tau)}$, where $P(\bar{c} \cap \bar{\tau})$ is the lowest possible probability, considering any couple of clusters and state-transitions.}

\section{ENVIRONMENTS} \label{sec:environments}

Real-life air/sea/road traffic regulations are often complex, and their mastery is a crucial aspect of orderly navigation. Many realistic settings have a number of exceptions that must be taken into consideration (e.g. ambulances are not subjected to some rules when in emergencies, sailing boats have different priorities if on wind power, etc). \edit{To evaluate the effect of XAER in a diverse configuration space of environments, we developed modular environments that allow us to systematically change its properties in evaluation. Our environments allow for agents to experience the same rules (our Easy, Medium, and Hard rulesets) in both discrete and continuous state-action spaces, and with frequent and sparse rewards. In these, the agent must understand the complex regulation governing the penalty system.} To implement our rulesets, we use \textit{cultures} \cite{Raymond2020Culture-BasedDeconfliction} \cite{Raymond2021AgreeResolution}: a mechanism to encode human rulesets as machine-compatible argumentation frameworks, imbued with fact-checking mechanisms. These can \edit{serve as explainer functions to} produce rule-based explanations from an agent's behaviour.

\begin{figure}[!htb]
	\centering
	\includegraphics[width=\columnwidth]{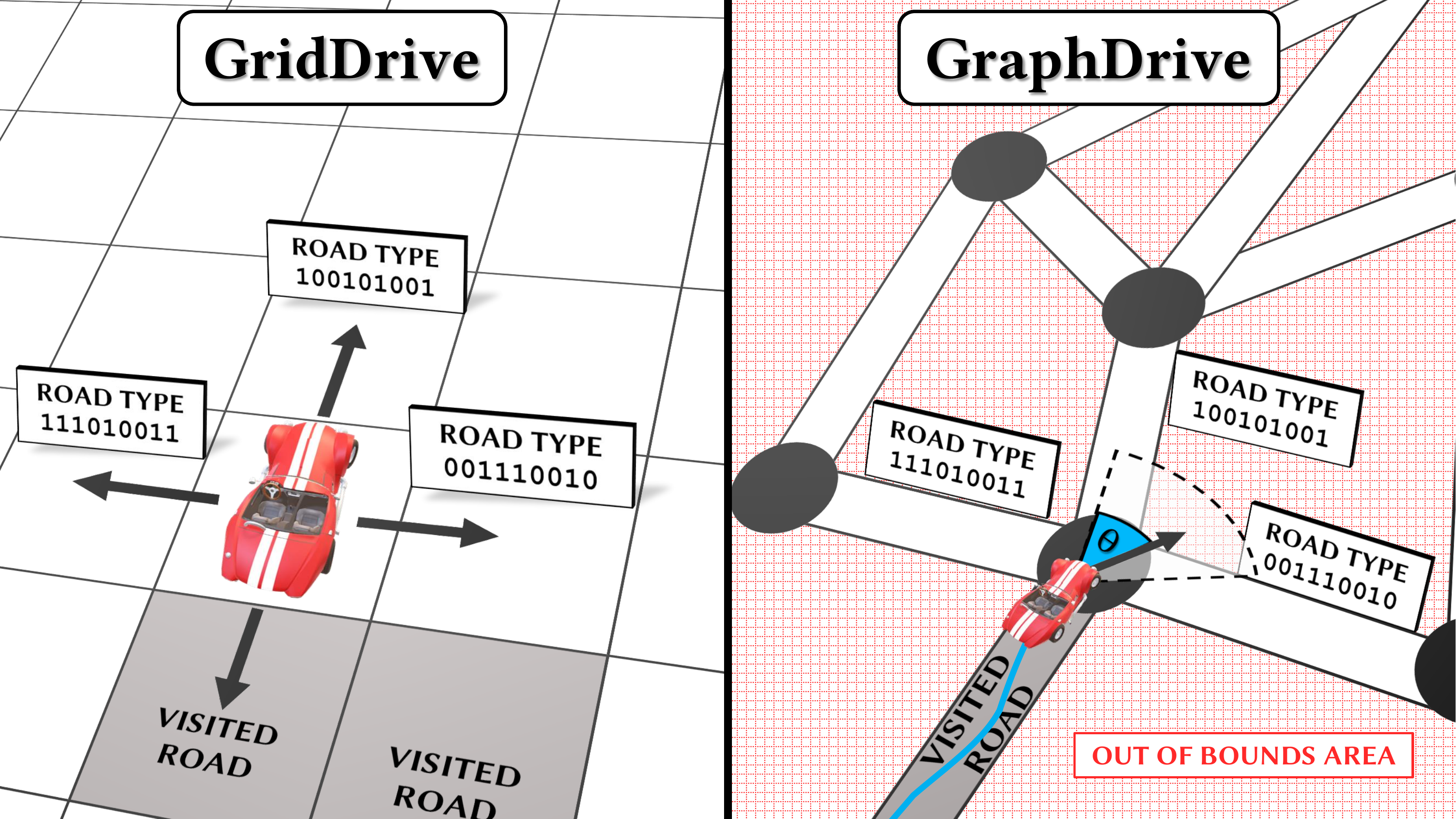}
	\caption{Diagrams representing our proposed GridDrive and GraphDrive environments. In GridDrive, the agent has a discrete action space and must observe the properties of neighbouring cells to make a decision that is compatible with the ruleset, choosing one direction and a fixed speed. GraphDrive is a harder environment, where the agent's action and observation spaces are continuous. In it, kinematics are taken into consideration and the agent must not only learn the rules governing penalties, but also to accelerate and steer without going off-road. The goal in both environments is to visit as many new roads as possible without infringing rules.}
	\label{fig:environments}
\end{figure}

The environments are:

\subsection {GridDrive - Discrete} A 15$\times$15 grid of cells, where every cell represents a different type of road (see Figure~\ref{fig:environments}, left), with base types (e.g. motorway, school road, city) combined with other modifiers (roadworks, accidents, weather). Each vehicle will have a set of properties that define which type of vehicle they are (emergency, civilian, worker, etc). Complex combinations of these properties will define a strict speed limit for each cell, according to the culture. 

\textit{Actions.} A sample $(d, s)$ in the action space consists of a direction $d \in \{N, S, E, W\}$ and a speed $s$ where $0 < s \leq 12$.

\textit{Observations.} A sample in the observation space is a tuple $(o_v, o_r, M, x, y)$ where $o_v$ denotes the concatenation of the vehicle's properties (including speed), $o_r$ is the concatenation of all neighbouring roads' properties, $M$ is a $15 \times 15 \times 2$ boolean matrix keeping track of visited cells, and $(x, y)$ represent the vehicle's current global coordinates. 

\textit{Rewards.} Let $0 < s' \leq 1$ denote the \textit{normalised} speed of the agent in that step. Rewards are given at every step, given the following criteria: 
\[
        \begin{cases} 
          -1\text{ (terminal)}, & \text{if breaking the speed regulation} \\
          0, & \text{if on previously-visited cell} \\
          s', & \text{otherwise (new cell, within speed limit)}
        \end{cases}
\]

\subsection {GraphDrive - Continuous} An Euclidean representation of a \textit{planar} graph with $n$ vertices and $m$ edges (see Figure~\ref{fig:environments}, right). The agent starts at the coordinates of one of those vertices and has to drive between vertices (called `junctions') in continuous space with Ackermann-based non-holonomic motion. Edges represent roads and are subjected to the same rules with properties to those seen in GridDrive plus a few extra rules to encourage the agent to stay close to the edges. The incentive is to drive as long as possible without committing speed infractions. In this setting, the agent \edit{must learn a control input that not only keeps the vehicle on the road, but also respects speed limits and restrictions that may vary on a case-by-case basis.} We test two variations of this environment: one with dense and another with sparse rewards.

\textit{Actions.} A sample $(\theta, a)$ in the action space consists of a steering angle $\theta$ where $-\frac{\pi}{4} \leq \theta \leq \frac{\pi}{4}$, and an acceleration $a$ where $-7 \leq a \leq 1$. Acceleration and deceleration ranges are chosen given road car standards (in $m/s^2$) \cite{Bokare2017Acceleration-DecelerationTypes} \cite{Fambro2000DriverSituations}.

\textit{Observations.} A sample in the observation space for GraphDrive is a tuple $(o_v, o_r, o_j)$, where $o_v$ denotes a concatenation of the vehicle's properties (car features, position, speed/angle, distance to path, junction status, number of visited junctions), $o_r$ is the concatenation of the properties of the closest road to the agent (likely to be the one the agent is driving on), and $o_j$ is the concatenation of the properties of roads connected to the next junction.

\textit{Rewards (dense version).} Let $0 < s \leq 1$ denote the \textit{normalised} speed of the agent in that frame, and let $n$ be the number of \textit{unique} junctions visited in the episode. Rewards are given at every frame, given the following criteria:
\[
        \begin{cases} 
          -1\text{ (terminal)}, & \text{if breaking the speed regulation} \\
          -1\text{ (terminal)}, & \text{if off-road or U-turning outside junction} \\
          0, & \text{if on junction or previously-visited road} \\
          s, & \text{otherwise (on road, within speed limit)}
        \end{cases}
\]

\textit{Rewards (sparse version).} In this version, the agent will get null (zero) reward when moving correctly. Positive rewards only appear when the agent manages to acquire a new junction. Therefore, the agent will have to drive entire roads correctly to get any positive reward. Rewards are given according to the following criteria:
\[
        \begin{cases} 
          -1\text{ (terminal)}, & \text{if breaking the speed regulation} \\
          -1\text{ (terminal)}, & \text{if off-road or U-turning outside junction} \\
          0, & \text{driving normally or on acquired junction} \\
          1, & \text{the instant a new junction is acquired}
        \end{cases}
\]

Every episode incurs in an initialisation of the grid or graph (for GridDrive or GraphDrive, respectively) with random roads, along with randomly-sampled agent properties. The agent is encouraged to drive for as long as possible until it either achieves a maximum number of steps or breaks a rule (terminal state). All environments will be instantiated in versions with 3 different cultures (rulesets), according to their levels of complexity: Easy, Medium, and Hard.

\begin{itemize}
    \item Easy: 3 properties (2 for roads, 1 for agents), 5 distinct explanations.%
    \item Medium: 7 properties (5 for roads, 2 for agents), 12 distinct explanations.%
    \item Hard: 15 properties (9 for roads, 6 for agents), 20 distinct explanations.%
\end{itemize}

\section{EXPERIMENTS}

In this section we describe our experimental setup and present results obtained in our proposed environments with XAER versus traditional PER. We trained 3 baseline agents with traditional PER (DQN/Rainbow, SAC, and TD3). For each of the 3 baseline algorithms, we train 3 XAER versions with different clustering strategies, using \texttt{HOW}, \texttt{WHY}, and \texttt{HOW+WHY} explanations (see Section~\ref{sec:xa}). Additionally, we show results for \texttt{HOW+WHY} explanations \textit{without} the simplicity heuristic (prioritised clustering) --- i.e. clusters are sampled uniformly. For a total of 12 XA agents, we call the XAER-equipped versions of DQN, SAC, and TD3 \textit{XADQN}, \textit{XASAC}, and \textit{XATD3}, respectively. DQN and XADQN agents are applied to GridDrive (discrete), whilst SAC, TD3, XASAC, and XATD3\footnote{Their implementations come from RLlib \cite{Liang2017RayLibrary}, an open-source library for RL agents. We developed the XARL Python library, which can be easily integrated in RLlib and provides XA facilities for obtaining XADQN, XATD3 and XASAC.} were trained separately on GraphDrive with dense and sparse rewards (continuous).

The neural network adopted for all the experiments is the default one implemented in the respective baselines (although better ones can be certainly devised), and it is characterised by fully connected layers of few units (e.g. 256) followed by the output layers for actors and/or critics, depending on the algorithm’s architecture. XAER methods introduce the cluster size proportion ($\xi$) hyperparameter. We perform ablation experiments to choose values of $\xi$, and arrive at $\xi=1$ for XADQN and XATD3, and $\xi=3$ for XASAC. We omit the detailed ablation study for brevity, but full plots and auxiliary results can be found in our GitHub page\footnote{\url{https://github.com/proroklab/xaer}}.

As the environments presented in Section \ref{sec:environments} have different levels of rule density/complexity, we are interested in observing if XAER exhibits superior performance compared to traditional PER in tasks that involve learning sophisticated and exception-heavy regulations. We trained all agents up to $4.0 \times 10^7$ steps \edit{sampled} on all environments \edit{for a total of approximately $1.6 \times 10^8$ training steps}. Our reported scores are obtained by segmenting the curve of mean episode rewards into 20 regions containing 5\% of steps each. We select the best region (highest median) for each agent to compare agents at their respective best performances. We report those medians in Table~\ref{tab:drive_results}, as well as the 25-75\% inter-quartile range for the selected region.

\begin{table*}[]
\caption{Median cumulative rewards after $4.0 \times 10^7$ steps for experiments on GridDrive, GraphDrive, and GraphDrive with sparse rewards (SR). Darker cells indicate better results in environment. Bold are best in row. IQR (25\%-75\%) in brackets.}
\label{tab:drive_results}
\begin{tabular}{cccccc}
\hline
\rowcolor[HTML]{C0C0C0} 
\multicolumn{1}{|c|}{\cellcolor[HTML]{FFFFD3}DQN/Rainbow}               & \multicolumn{1}{c|}{\cellcolor[HTML]{C0C0C0}\textit{Baseline}}             & \multicolumn{1}{c|}{\cellcolor[HTML]{C0C0C0}\textit{XADQN-HOW}}            & \multicolumn{1}{c|}{\cellcolor[HTML]{C0C0C0}\textit{XADQN-WHY}}               & \multicolumn{1}{c|}{\cellcolor[HTML]{C0C0C0}\textit{XADQN-HOW+WHY}}           & \multicolumn{1}{c|}{\cellcolor[HTML]{C0C0C0}\textit{\begin{tabular}[c]{@{}c@{}}XADQN-HOW+WHY\\ sans simplicity\end{tabular}}} \\ \hline
\multicolumn{1}{|c|}{\cellcolor[HTML]{DFDFDF}Grid Easy}         & \multicolumn{1}{c|}{\cellcolor[HTML]{BDD7EE}\textbf{17.13 (16.02-18.03)}} & \multicolumn{1}{c|}{\cellcolor[HTML]{E7EFF4}14.84 (12.88-15.88)}      & \multicolumn{1}{c|}{\cellcolor[HTML]{F2F2F2}13.68 (11.73-15.29)}             & \multicolumn{1}{c|}{\cellcolor[HTML]{E8EFF4}14.7 (13.08-15.91)}             & \multicolumn{1}{c|}{\cellcolor[HTML]{E6EEF4}14.88 (13.33-16.04)}                                                             \\ \hline
\multicolumn{1}{|c|}{\cellcolor[HTML]{DFDFDF}Grid Medium}       & \multicolumn{1}{c|}{\cellcolor[HTML]{EFF1F2}7.99 (7.05-8.9)}             & \multicolumn{1}{c|}{\cellcolor[HTML]{F2F2F2}7.59 (6.7-8.59)}         & \multicolumn{1}{c|}{\cellcolor[HTML]{EEF1F2}8.06 (7.17-9.09)}                & \multicolumn{1}{c|}{\cellcolor[HTML]{BDD7EE}\textbf{11.62 (10.48-12.66)}}    & \multicolumn{1}{c|}{\cellcolor[HTML]{E4EEF5}9.21 (7.79-10.46)}                                                                \\ \hline
\multicolumn{1}{|c|}{\cellcolor[HTML]{DFDFDF}Grid Hard}         & \multicolumn{1}{c|}{\cellcolor[HTML]{E2EDF5}1.99 (1.74-2.24)}             & \multicolumn{1}{c|}{\cellcolor[HTML]{E2EDF5}1.97 (1.72-2.24)}         & \multicolumn{1}{c|}{\cellcolor[HTML]{E6EEF5}1.75 (1.51-2.03)}                & \multicolumn{1}{c|}{\cellcolor[HTML]{BDD7EE}\textbf{3.14 (2.73-3.62)}}       & \multicolumn{1}{c|}{\cellcolor[HTML]{F2F2F2}0.95 (0.8 - 1.14)}                                                                 \\ \hline
                                                                &                                                                            &                                                                        &                                                                               &                                                                               &                                                                                                                               \\ \hline
\rowcolor[HTML]{C0C0C0} 
\multicolumn{1}{|c|}{\cellcolor[HTML]{FFFFD3}TD3}               & \multicolumn{1}{c|}{\cellcolor[HTML]{C0C0C0}\textit{Baseline}}             & \multicolumn{1}{c|}{\cellcolor[HTML]{C0C0C0}\textit{XATD3-HOW}}        & \multicolumn{1}{c|}{\cellcolor[HTML]{C0C0C0}\textit{XATD3-WHY}}               & \multicolumn{1}{c|}{\cellcolor[HTML]{C0C0C0}\textit{XATD3-HOW+WHY}}           & \multicolumn{1}{c|}{\cellcolor[HTML]{C0C0C0}\textit{\begin{tabular}[c]{@{}c@{}}XATD3-HOW+WHY\\ sans simplicity\end{tabular}}} \\ \hline
\multicolumn{1}{|c|}{\cellcolor[HTML]{DFDFDF}Graph Easy}        & \multicolumn{1}{c|}{\cellcolor[HTML]{D3E5F5}75.48 (68.09-80.85)}          & \multicolumn{1}{c|}{\cellcolor[HTML]{F2F2F2}0.0 (-0.02-0.02)}       &  \multicolumn{1}{c|}{\cellcolor[HTML]{C9DFF2}88.75 (83.29-94.44)}          &
\multicolumn{1}{c|}{\cellcolor[HTML]{BDD7EE}\textbf{103.72 (98.64-107.03)}} &\multicolumn{1}{c|}{\cellcolor[HTML]{CDE1F3}84.23 (79.28-89.2)}                                                            \\ \hline
\multicolumn{1}{|c|}{\cellcolor[HTML]{DFDFDF}Graph Medium}      & \multicolumn{1}{c|}{\cellcolor[HTML]{C9DFF2}75.48 (68.09-80.85)}          & \multicolumn{1}{c|}{\cellcolor[HTML]{F2F2F2}41.31 (33.24-47.49)}      & \multicolumn{1}{c|}{\cellcolor[HTML]{DBEAF7}64.8 (59.44-69.47)}          & \multicolumn{1}{c|}{\cellcolor[HTML]{BDD7EE}\textbf{78.34 (73.21-83.07)}} & \multicolumn{1}{c|}{\cellcolor[HTML]{D1E4F4}69.36 (61.01-77.58)}                                                            \\ \hline
\multicolumn{1}{|c|}{\cellcolor[HTML]{DFDFDF}Graph Hard}        & \multicolumn{1}{c|}{\cellcolor[HTML]{F2F2F2}-0.01 (-0.03-0.0)}          & \multicolumn{1}{c|}{\cellcolor[HTML]{F2F2F2}-0.01 (-0.03-0.0)}      & \multicolumn{1}{c|}{\cellcolor[HTML]{BDD7EE}\textbf{20.65 (18.9-22.4)}}    & \multicolumn{1}{c|}{\cellcolor[HTML]{D5E6F5}14.54 (13.17-16.12)}             & \multicolumn{1}{c|}{\cellcolor[HTML]{E1EDF6}10.31 (8.84-11.68)}                                                             \\ \hline
\multicolumn{1}{|c|}{\cellcolor[HTML]{DFDFDF}Graph Easy (SR)}   & \multicolumn{1}{c|}{\cellcolor[HTML]{BDD7EE}\textbf{2.65 (2.28-2.93)}}             & \multicolumn{1}{c|}{\cellcolor[HTML]{F2F2F2}-0.04 (-0.06-(-0.02))}          & \multicolumn{1}{c|}{\cellcolor[HTML]{BFD8EF}2.61 (2.43-2.75)}       & \multicolumn{1}{c|}{\cellcolor[HTML]{C0D9EF}2.55 (2.42-2.66)}                & \multicolumn{1}{c|}{\cellcolor[HTML]{C3DBF0}2.47 (2.34-2.62)}                                                                 \\ \hline
\multicolumn{1}{|c|}{\cellcolor[HTML]{DFDFDF}Graph Medium (SR)} & \multicolumn{1}{c|}{\cellcolor[HTML]{EEF1F3}0.34 (-1.0-0.97)}             & \multicolumn{1}{c|}{\cellcolor[HTML]{F2F2F2}-0.04 (-0.05-(-0.03))}  & \multicolumn{1}{c|}{\cellcolor[HTML]{C4DBF0}2.54 (2.3-2.79)}                    & \multicolumn{1}{c|}{\cellcolor[HTML]{BDD7EE}\textbf{2.75 (2.58-2.96)}}                 & \multicolumn{1}{c|}{\cellcolor[HTML]{D8E8F6}1.84 (1.47-2.0)}                                                                 \\ \hline
\multicolumn{1}{|c|}{\cellcolor[HTML]{DFDFDF}Graph Hard (SR)}   & \multicolumn{1}{c|}{\cellcolor[HTML]{F2F2F2}-0.03 (-0.05-(-0.02))}           & \multicolumn{1}{c|}{\cellcolor[HTML]{F2F2F2}-0.04 (-0.05-(-0.03))} & \multicolumn{1}{c|}{\cellcolor[HTML]{F2F2F2}-0.04 (-0.06-(-0.03))}                 & \multicolumn{1}{c|}{\cellcolor[HTML]{F2F2F2}-0.05 (-0.06-(-0.03))}              & \multicolumn{1}{c|}{\cellcolor[HTML]{F2F2F2}-0.05 (-0.6-(-0.04))}                                                                 \\ \hline
                                                                &                                                                            &                                                                        &                                                                               &                                                                               &                                                                                                                               \\ \hline
\rowcolor[HTML]{C0C0C0} 
\multicolumn{1}{|c|}{\cellcolor[HTML]{FFFFD3}SAC}               & \multicolumn{1}{c|}{\cellcolor[HTML]{C0C0C0}\textit{Baseline}}             & \multicolumn{1}{c|}{\cellcolor[HTML]{C0C0C0}\textit{XASAC-HOW}}        & \multicolumn{1}{c|}{\cellcolor[HTML]{C0C0C0}\textit{XASAC-WHY}}               & \multicolumn{1}{c|}{\cellcolor[HTML]{C0C0C0}\textit{XASAC-HOW+WHY}}           & \multicolumn{1}{c|}{\cellcolor[HTML]{C0C0C0}\textit{\begin{tabular}[c]{@{}c@{}}XASAC-HOW+WHY\\ sans simplicity\end{tabular}}} \\ \hline
\multicolumn{1}{|c|}{\cellcolor[HTML]{DFDFDF}Graph Easy}        & \multicolumn{1}{c|}{\cellcolor[HTML]{F2F2F2}65.9 (59.04-72.94)}          & \multicolumn{1}{c|}{\cellcolor[HTML]{ECF0F3}79.46 (71.72-88.46)}      & \multicolumn{1}{c|}{\cellcolor[HTML]{C0D9EF}138.81 (133.0-144.05)}            & \multicolumn{1}{c|}{\cellcolor[HTML]{BDD7EE}\textbf{141.11 (136.45-145.87)}}            & \multicolumn{1}{c|}{\cellcolor[HTML]{D8E8F6}116.39 (110.36-120.9)}                                                   \\ \hline
\multicolumn{1}{|c|}{\cellcolor[HTML]{DFDFDF}Graph Medium}      & \multicolumn{1}{c|}{\cellcolor[HTML]{F2F2F2}65.78 (58.43-71.92)}             & \multicolumn{1}{c|}{\cellcolor[HTML]{EAF0F3}76.61 (69.64-83.69)}         & \multicolumn{1}{c|}{\cellcolor[HTML]{BDD7EE}\textbf{112.16 (105.87-119.1)}}    & \multicolumn{1}{c|}{\cellcolor[HTML]{BFD8EF}111.4 (106.72-116.11)}             & \multicolumn{1}{c|}{\cellcolor[HTML]{D6E7F5}97.81 (92.91-103.1)}                                                              \\ \hline
\multicolumn{1}{|c|}{\cellcolor[HTML]{DFDFDF}Graph Hard}        & \multicolumn{1}{c|}{\cellcolor[HTML]{DDEBF7}26.85 (24.43-28.66)}            & \multicolumn{1}{c|}{\cellcolor[HTML]{E6EEF4}22.92 (20.61-25.03)}      &  \multicolumn{1}{c|}{\cellcolor[HTML]{C0D9EF}32.14 (29.93-34.49)}             &
\multicolumn{1}{c|}{\cellcolor[HTML]{BDD7EE}\textbf{32.58 (30.41-34.69)}}    &\multicolumn{1}{c|}{\cellcolor[HTML]{F2F2F2}17.85 (13.82-20.56)}                                                                 \\ \hline
\multicolumn{1}{|c|}{\cellcolor[HTML]{DFDFDF}Graph Easy (SR)}   & \multicolumn{1}{c|}{\cellcolor[HTML]{DFECF6}3.57 (3.19-4.01)}              & \multicolumn{1}{c|}{\cellcolor[HTML]{E5EEF5}3.07 (2.82-3.21)}         & \multicolumn{1}{c|}{\cellcolor[HTML]{BED8EF}4.82 (4.64-4.98)}                & \multicolumn{1}{c|}{\cellcolor[HTML]{BDD7EE}\textbf{4.83 (4.58-5.09)}}       & \multicolumn{1}{c|}{\cellcolor[HTML]{F2F2F2}2.01 (1.8-2.19)}                                                                 \\ \hline
\multicolumn{1}{|c|}{\cellcolor[HTML]{DFDFDF}Graph Medium (SR)} & \multicolumn{1}{c|}{\cellcolor[HTML]{C9DFF2}2.61 (2.26-2.85)}           & \multicolumn{1}{c|}{\cellcolor[HTML]{BDD7EE}\textbf{2.66 (2.26-2.98)}}       & \multicolumn{1}{c|}{\cellcolor[HTML]{C2DAF0}2.64 (2.53-2.75)}       & \multicolumn{1}{c|}{\cellcolor[HTML]{E2EDF5}2.47 (2.33-2.55)}                & \multicolumn{1}{c|}{\cellcolor[HTML]{F2F2F2}2.31 (2.17-2.45)}                                                                 \\ \hline
\rowcolor[HTML]{F2F2F2} 
\multicolumn{1}{|c|}{\cellcolor[HTML]{DFDFDF}Graph Hard (SR)\footnotemark}   & \multicolumn{1}{c|}{\cellcolor[HTML]{D0E3F4}1.15 (1.03-1.27)}           & \multicolumn{1}{c|}{\cellcolor[HTML]{BDD7EE}\textbf{1.53 (1.37-1.63)}}       & \multicolumn{1}{c|}{\cellcolor[HTML]{D2E4F4}1.11 (1.01-1.23)}              & \multicolumn{1}{c|}{\cellcolor[HTML]{F2F2F2}-0.09 (-0.12-(-0.07))}              & \multicolumn{1}{c|}{\cellcolor[HTML]{DFECF6}0.81 (0.65-0.94)}                                                              \\  \hline
\end{tabular}
\end{table*}

Results in Table~\ref{tab:drive_results} show that across all tasks and methods, XAER versions only lose to the PER baseline against DQN/Rainbow in GridDrive Easy, by 0.4\%. For GridDrive Medium and Hard, XADQN with \texttt{HOW+WHY} explanations exhibit significantly higher performance (57\% and 81\%, respectively). \texttt{WHY} and \texttt{HOW+WHY} exhibit similar performance in GraphDrive, being bested by \texttt{HOW} in Medium and Hard \textit{Sparse} cases only. Although \texttt{HOW+WHY} explanations have consistently good results across environments, the version without the simplicity heuristic exhibited consistently inferior results. Neither baseline SAC or TD3 managed to learn a policy in GraphDrive Hard Sparse (our hardest environment). XATD3 also failed to learn a policy in this environment, but XASAC was able to achieve positive results.

\section{DISCUSSION AND CONCLUSION}

Our results indicate a significant benefit achieved via explanation-aware experience replay.  In one case (TD3 Hard), endowing an agent with XAER \textit{enabled} an agent to learn altogether where it would otherwise fail entirely. XAER allowed agents to learn in Medium and Hard difficulty settings, obtaining significantly higher rewards whilst having the same hyper-parameters and number of learning steps.

The choice of explanation type also affected results: when superior, \texttt{HOW+WHY} explanations exhibited larger margins of improvement over other XAER methods. In other cases, when bested by \texttt{WHY} explanations, the former maintained very close results, thus achieving consistently satisfactory results in most cases. Also importantly, although \texttt{HOW} explanations exhibited lower performance than other XAER counterparts in most environments, it is worth noting that \texttt{HOW} explanations do not require an explainer and could in theory be used in any environment. The consistency of \texttt{HOW+WHY} results suggests that the act of explaining may involve answering more archetypal questions, not just causal ones, as hypothesised also in \cite{sovrano2021philosophy}.

The frequency and magnitude of rewards is an important factor to be considered in XAER clustering. When negative rewards are more frequent (with similar magnitude to positive rewards), and there are more negative than positive clusters, oversampling may cause the agent to tackle situations with negative rewards more frequently, preventing it to maximise cumulative rewards. This effect can be particularly pronounced with very sparse rewards, such as the ones seen in the sparse version of GraphDrive.

Intuitively, this is akin to the notion that if there are few opportunities to explain, one must choose their explanations well. The notion of \textit{explanation engineering} surfaces as a mechanism to orient the learning agent through means of selecting which experiences (and explanations) are more important to the task at hand, by means of \textit{abstractions}. 
Being explainable by design, explanation engineering can be an intuitive and \textit{semantically-grounded} alternative to reward engineering, as the \textit{meaning} of the rewards matter just as their magnitude. 
A few examples include increasing the number of positive clusters, or organise clusters hierarchically.

With regards to \textit{relevance}, if the cumulative priority of the state-transitions of a whole cluster is low, it may indicate that the agent has already learned to handle the task represented by the cluster, so it may not need it as an explanation (thus being less relevant). Oppositely, if the cumulative priority is high, it could indicate a further need for additional explanations. The cluster might be representing either non-generic or generic tasks. If the agent needs explanations for a generic task, it should also need them for a non-generic task. In that case, the generic task is prioritised over the non-generic. The benefits of inter-cluster prioritisation (\textit{simplicity}) are higher in environments with harder rulesets, and proportional to the complexity of the culture \cite{Raymond2020Culture-BasedDeconfliction}. This suggests that uniformly selecting an explanation type to replay is less beneficial than selecting the simplest and most relevant explanation.

This work foments diverse avenues for further investigation. For one, further experiments could include the development of explainer functions to evaluate the performance of \texttt{WHY} explanations in popular benchmarks. Additionally, future work may observe the effect of XAER with on-policy algorithms, such as PPO. And lastly, the illocutionary effect of explanations deriving from further archetypal questions \cite{sovrano2021philosophy} (i.e. \texttt{WHAT}, \texttt{WHERE}, \texttt{WHEN}) could be explored in advanced explanation engineering for experience clustering.

\addtolength{\textheight}{-0cm}   %

\section*{ACKNOWLEDGEMENT}
We would like to thank the anonymous reviewers for their constructive remarks and suggestions. We thank our colleagues Nikhil Churamani, Guillaume Sartoretti and Yuhong Cao for providing commentary in early design discussions, as well as reviewing our pre-submission manuscript.

\bibliographystyle{plain}

\begin{thebibliography}{10}

\bibitem{Aamodt1994Case-BasedApproaches}
Agnar Aamodt and Enric Plaza.
\newblock {Case-Based Reasoning: Foundational Issues, Methodological
  Variations, and System Approaches}.
\newblock {\em AI Communications}, 7:39--59, 1994.

\bibitem{achinstein1983nature}
Peter Achinstein.
\newblock {\em The nature of explanation}.
\newblock Oxford University Press on Demand, 1983.

\bibitem{BarredoArrieta2020ExplainableAI}
Alejandro Barredo~Arrieta, Natalia D{\'{i}}az-Rodr{\'{i}}guez, Javier Del~Ser,
  Adrien Bennetot, Siham Tabik, Alberto Barbado, Salvador Garcia, Sergio
  Gil-Lopez, Daniel Molina, Richard Benjamins, Raja Chatila, and Francisco
  Herrera.
\newblock {Explainable Artificial Intelligence (XAI): Concepts, taxonomies,
  opportunities and challenges toward responsible AI}.
\newblock {\em Information Fusion}, 58:82--115, 2020.

\bibitem{Berkenkamp2017SafeGuarantees}
Felix Berkenkamp, Matteo Turchetta, Angela Schoellig, and Andreas Krause.
\newblock {Safe Model-based Reinforcement Learning with Stability Guarantees}.
\newblock In I~Guyon, U~V Luxburg, S~Bengio, H~Wallach, R~Fergus,
  S~Vishwanathan, and R~Garnett, editors, {\em Advances in Neural Information
  Processing Systems}, volume~30. Curran Associates, Inc., 2017.

\bibitem{Blumer1987OccamsRazor}
Anselm Blumer, Andrzej Ehrenfeucht, David Haussler, and Manfred~K Warmuth.
\newblock {Occam's razor}.
\newblock {\em Information processing letters}, 24(6):377--380, 1987.

\bibitem{Bokare2017Acceleration-DecelerationTypes}
P~S Bokare and A~K Maurya.
\newblock {Acceleration-Deceleration Behaviour of Various Vehicle Types}.
\newblock {\em Transportation Research Procedia}, 25:4733--4749, 2017.

\bibitem{Branting1991BuildingCases}
L.Karl Branting.
\newblock {Building explanations from rules and structured cases}.
\newblock {\em International Journal of Man-Machine Studies}, 34(6):797--837,
  1991.

\bibitem{chow2018lyapunov}
Yinlam Chow, Ofir Nachum, Edgar Duenez-Guzman, and Mohammad Ghavamzadeh.
\newblock A lyapunov-based approach to safe reinforcement learning.
\newblock {\em arXiv preprint arXiv:1805.07708}, 2018.

\bibitem{dietterich1997explanation}
Thomas~G Dietterich and Nicholas~S Flann.
\newblock Explanation-based learning and reinforcement learning: A unified
  view.
\newblock {\em Machine Learning}, 28(2):169--210, 1997.

\bibitem{Fambro2000DriverSituations}
Daniel~B Fambro, Rodger~J Koppa, Dale~L Picha, and Kay Fitzpatrick.
\newblock {Driver Braking Performance in Stopping Sight Distance Situations}.
\newblock {\em Transportation Research Record}, 1701(1):9--16, 1 2000.

\bibitem{Fujimoto2018AddressingMethods}
Scott Fujimoto, Herke van Hoof, and David Meger.
\newblock {Addressing Function Approximation Error in Actor-Critic Methods}.
\newblock In Jennifer Dy and Andreas Krause, editors, {\em Proceedings of the
  35th International Conference on Machine Learning}, volume~80 of {\em
  Proceedings of Machine Learning Research}, pages 1587--1596. PMLR, 4 2018.

\bibitem{garcia2015comprehensive}
Javier Garc{\i}a and Fernando Fern{\'a}ndez.
\newblock A comprehensive survey on safe reinforcement learning.
\newblock {\em Journal of Machine Learning Research}, 16(1):1437--1480, 2015.

\bibitem{Haarnoja2018SoftActor}
Tuomas Haarnoja, Aurick Zhou, Pieter Abbeel, and Sergey Levine.
\newblock {Soft Actor-Critic: Off-Policy Maximum Entropy Deep Reinforcement
  Learning with a Stochastic Actor}.
\newblock In Jennifer Dy and Andreas Krause, editors, {\em Proceedings of the
  35th International Conference on Machine Learning}, volume~80 of {\em
  Proceedings of Machine Learning Research}, pages 1861--1870,
  Stockholmsm{\"{a}}ssan, Stockholm Sweden, 3 2018. PMLR.

\bibitem{holland1989induction}
John~H Holland, Keith~J Holyoak, Richard~E Nisbett, and Paul~R Thagard.
\newblock {\em Induction: Processes of inference, learning, and discovery}.
\newblock MIT press, 1989.

\bibitem{hussein2017imitation}
Ahmed Hussein, Mohamed~Medhat Gaber, Eyad Elyan, and Chrisina Jayne.
\newblock Imitation learning: A survey of learning methods.
\newblock {\em ACM Computing Surveys (CSUR)}, 50(2):1--35, 2017.

\bibitem{johnson2019simplicity}
Samuel~GB Johnson, JJ~Valenti, and Frank~C Keil.
\newblock Simplicity and complexity preferences in causal explanation: An
  opponent heuristic account.
\newblock {\em Cognitive psychology}, 113:101222, 2019.

\bibitem{Kaiser2019Model-basedAtari}
Lukasz Kaiser, Mohammad Babaeizadeh, Piotr Milos, Blazej Osinski, Roy~H
  Campbell, Konrad Czechowski, Dumitru Erhan, Chelsea Finn, Piotr Kozakowski,
  Sergey Levine, and {others}.
\newblock {Model-based reinforcement learning for atari}.
\newblock {\em arXiv preprint arXiv:1903.00374}, 2019.

\bibitem{Kiran2021DeepSurvey}
B~Ravi Kiran, Ibrahim Sobh, Victor Talpaert, Patrick Mannion, Ahmad A~Al
  Sallab, Senthil Yogamani, and Patrick P{\'{e}}rez.
\newblock {Deep Reinforcement Learning for Autonomous Driving: A Survey}.
\newblock {\em IEEE Transactions on Intelligent Transportation Systems}, pages
  1--18, 2021.

\bibitem{Li2021RevisitingPerspective}
Ang~A Li, Zongqing Lu, and Chenglin Miao.
\newblock {Revisiting Prioritized Experience Replay: A Value Perspective}.
\newblock {\em arXiv preprint arXiv:2102.03261}, 2021.

\bibitem{Li2019UrbanLearning}
Changjian Li and Krzysztof Czarnecki.
\newblock {Urban Driving with Multi-Objective Deep Reinforcement Learning}.
\newblock In {\em Proceedings of the 18th International Conference on
  Autonomous Agents and MultiAgent Systems}, AAMAS '19, pages 359--367,
  Richland, SC, 2019. International Foundation for Autonomous Agents and
  Multiagent Systems.

\bibitem{Liang2017RayLibrary}
Eric Liang, Richard Liaw, Robert Nishihara, Philipp Moritz, Roy Fox, Joseph
  Gonzalez, Ken Goldberg, and Ion Stoica.
\newblock {Ray rllib: A composable and scalable reinforcement learning
  library}.
\newblock {\em arXiv preprint arXiv:1712.09381}, page~85, 2017.

\bibitem{mayes2005theories}
GR~Mayes.
\newblock Theories of explanation. the internet encyclopedia of philosophy,
  2005.

\bibitem{Mnih2015Human-levelLearning}
Volodymyr Mnih, Koray Kavukcuoglu, David Silver, Andrei~A Rusu, Joel Veness,
  Marc~G Bellemare, Alex Graves, Martin Riedmiller, Andreas~K Fidjeland, Georg
  Ostrovski, Stig Petersen, Charles Beattie, Amir Sadik, Ioannis Antonoglou,
  Helen King, Dharshan Kumaran, Daan Wierstra, Shane Legg, and Demis Hassabis.
\newblock {Human-level control through deep reinforcement learning}.
\newblock {\em Nature}, 518(7540):529--533, 2015.

\bibitem{nielsen1994enhancing}
Jakob Nielsen.
\newblock Enhancing the explanatory power of usability heuristics.
\newblock In {\em Proceedings of the SIGCHI conference on Human Factors in
  Computing Systems}, pages 152--158, 1994.

\bibitem{Raymond2020Culture-BasedDeconfliction}
Alex Raymond, Hatice Gunes, and Amanda Prorok.
\newblock {Culture-Based Explainable Human-Agent Deconfliction}.
\newblock In {\em Proceedings of the 19th International Conference on
  Autonomous Agents and MultiAgent Systems}, AAMAS ’20, pages 1107--1115,
  Richland, SC, 2020. International Foundation for Autonomous Agents and
  Multiagent Systems.

\bibitem{Raymond2021AgreeResolution}
Alex Raymond, Matthew Malencia, Guilherme Paulino-Passos, and Amanda Prorok.
\newblock {Agree to Disagree: Subjective Fairness in Privacy-Restricted
  Decentralised Conflict Resolution}.
\newblock {\em arXiv preprint arXiv:2107.00032}, 2021.

\bibitem{ren2018self}
Zhipeng Ren, Daoyi Dong, Huaxiong Li, and Chunlin Chen.
\newblock Self-paced prioritized curriculum learning with coverage penalty in
  deep reinforcement learning.
\newblock {\em IEEE transactions on neural networks and learning systems},
  29(6):2216--2226, 2018.

\bibitem{Rong2020SafeDriving}
Jikun Rong and Nan Luan.
\newblock {Safe Reinforcement Learning with Policy-Guided Planning for
  Autonomous Driving}.
\newblock In {\em 2020 IEEE International Conference on Mechatronics and
  Automation (ICMA)}, pages 320--326, 2020.

\bibitem{Schaul2015PrioritizedReplay}
Tom Schaul, John Quan, Ioannis Antonoglou, and David Silver.
\newblock {Prioritized experience replay}.
\newblock {\em arXiv preprint arXiv:1511.05952}, 2015.

\bibitem{sovrano2019combining}
Francesco Sovrano.
\newblock Combining experience replay with exploration by random network
  distillation.
\newblock In {\em 2019 IEEE Conference on Games (CoG)}, pages 1--8. IEEE, 2019.

\bibitem{sovrano2021philosophy}
Francesco Sovrano and Fabio Vitali.
\newblock From philosophy to interfaces: an explanatory method and a tool based
  on achinstein's theory of explanation.
\newblock In {\em Proceedings of the 26th International Conference on
  Intelligent User Interfaces}, 2021.

\bibitem{sun2020attentive}
Peiquan Sun, Wengang Zhou, and Houqiang Li.
\newblock Attentive experience replay.
\newblock In {\em Proceedings of the AAAI Conference on Artificial
  Intelligence}, volume~34, pages 5900--5907, 2020.

\bibitem{yin2017knowledge}
Haiyan Yin and Sinno Pan.
\newblock Knowledge transfer for deep reinforcement learning with hierarchical
  experience replay.
\newblock In {\em Proceedings of the AAAI Conference on Artificial
  Intelligence}, volume~31, 2017.

\bibitem{zha2019experience}
Daochen Zha, Kwei-Herng Lai, Kaixiong Zhou, and Xia Hu.
\newblock Experience replay optimization.
\newblock {\em arXiv preprint arXiv:1906.08387}, 2019.

\end{thebibliography}

\footnotetext[5]{All agents failed to learn a policy and thus we do not highlight results.}

\end{document}